\documentclass{article} 
\usepackage{iclr2020_conference,times}

\usepackage{amsmath,amsfonts,bm}

\def\eqref#1{equation~\ref{#1}}

\def\1{\bm{1}}

\DeclareMathAlphabet{\mathsfit}{\encodingdefault}{\sfdefault}{m}{sl}
\SetMathAlphabet{\mathsfit}{bold}{\encodingdefault}{\sfdefault}{bx}{n}

\usepackage{hyperref}
\usepackage{url}
\usepackage{caption}
\usepackage{subfigure}
\usepackage{graphicx}
\usepackage{algorithm}
\usepackage{algorithmic}
\usepackage{wrapfig}
\iclrfinalcopy
\title{MGHRL: Meta Goal-generation for \\ Hierarchical Reinforcement Learning}

\author{Haotian Fu$^1$, Hongyao Tang$^1$, Jianye Hao$^1$, Wulong Liu$^2$, Chen Chen$^2$ \\
\{haotianfu, bluecontra, jianye.hao\}@tju.edu.cn, \{liuwulong, chenchen9\}@huawei.com\\
$^1$Tianjin University, $^2$Noah's Ark Lab, Huawei 
\And
}

\begin{document}
	
	\maketitle
	
	\begin{abstract}
		Most meta reinforcement learning (meta-RL) methods learn to adapt to new tasks by directly optimizing the parameters of policies over primitive action space. Such algorithms work well in tasks with relatively slight difference. However, when the task distribution becomes wider, it would be quite inefficient to directly learn such a meta-policy. In this paper, we propose a new meta-RL algorithm called Meta Goal-generation for Hierarchical RL (MGHRL). Instead of directly generating policies over primitive action space for new tasks, MGHRL learns to generate high-level meta strategies over subgoals given past experience and leaves the rest of how to achieve subgoals as independent RL subtasks. Our empirical results on several challenging simulated robotics environments show that our method enables more efficient and generalized meta-learning from past experience.
	\end{abstract}
	
	\section{Introduction}

	Human intelligence is remarkable for their fast adaptation to many new situations using the knowledge learned from past experience. However, agents trained by conventional Deep Reinforcement Learning (DRL) methods~\citep{DBLP:journals/nature/MnihKSRVBGRFOPB15, DBLP:journals/jmlr/LevineFDA16,DBLP:conf/iclr/2016} can only learn one separate policy per task, failing to generalize to new tasks without additional large amount of training data. Meta reinforcement learning~\citep{DBLP:conf/icml/FinnAL17,DBLP:conf/iclr/MishraR0A18,DBLP:journals/corr/DuanSCBSA16} addresses such problems by learning how to learn. Given a number of tasks with similar structures, meta-RL methods enable agents learn such structure from previous experience on many tasks. Thus when encountering a new task, agents can quickly adapt to it with only a small amount of interactions.
	
	Most current meta-RL methods leverage experience from previous tasks to adapt to new tasks by directly learning the policy parameters over primitive action space~\citep{DBLP:conf/icml/FinnAL17,DBLP:conf/icml/RakellyZFLQ19}. Such approaches suffer from two problems: (i) For complex tasks which require sophisticated control strategies, it would be quite inefficient to directly learn such policy with one nonlinear function approximator and the adaptation to new tasks is prone to be inaccurate. This problem can become more severe in sparse reward settings. (ii) Current meta-RL methods focus on tasks with narrow distribution, how to generalize to new tasks with much more difference remains a problem.

	To tackle such problems, we propose a new hierarchical meta-RL method that meta-learns high-level goal generation and leaves the learning of low-level policy for independent RL. Intuitively, this is quite similar to how a human being behaves: we usually transfer the overall understanding of similar tasks rather than remember specific actions. Our meta goal-generation framework is built on top of the architecture of PEARL~\citep{DBLP:conf/icml/RakellyZFLQ19} and a two level hierarchy inspired by HAC~\citep{DBLP:conf/iclr/LevyKPS19}. Our evaluation on several simulated robotics tasks~\citep{DBLP:journals/corr/abs-1802-09464} as well as some human-engineered wider-distribution tasks shows the superiority of MGHRL to state-of-the-art meta-RL method.

	\section{Preliminaries}
	\label{background}
	
	In our meta learning scenario, we assume a distribution of tasks $p(\tau)$ that we want our model to adapt to. Each task correspond to a different Markov Decision Process (MDP), $M_{i}=\{S,A,T_{i},R_{i}\}$, with state space $S$, action space $A$, transition distribution $T_{i}$, and reward function $R_{i}$. We assume that the transitions and reward function vary across tasks. Meta-RL aims to learn a policy that can adapt to maximize the expected reward for novel tasks from $p(\tau)$ as efficiently as possible. 
	
	PEARL~\citep{DBLP:conf/icml/RakellyZFLQ19} is an off-policy meta-reinforcement learning method that drastically improves sample efficiency comparing to previous meta-RL algorithms. The meta-training process of PEARL learns a policy that adapts to the task at hand by conditioning the history of past transitions, which we refer to as context $c$. Specifically, for the $i$th transition in task $\tau$, $c_{i}^{\tau}=(s_{i},a_{i},r_{i},s'_{i})$. PEARL leverages an inference network $q_{\phi}(z|c)$ and outputs probabilistic latent variable $z$. The parameters of $q(z|c)$ are optimized jointly with the parameters of the actor $\pi_{\theta}(a|s,z)$ and critic $Q_{\theta}^{h}(s,a,z)$, using the reparametrization trick~\citep{DBLP:journals/corr/KingmaW13} to compute gradients for parameters of $q_{\phi}(z|c)$ through sampled probabilistic latent variable $z$.

	\section{Algorithm}
	\label{alg}

	\subsection{Two-level hierarchy}
	
    \begin{wrapfigure}{r}{5.4cm}
        \vspace{-0.5cm}
        \centering
		\includegraphics[width=5.4cm]{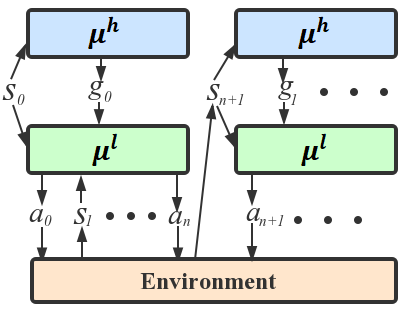}
		\caption{Two level Hierarchy}
		\vspace{-0.5cm}
		\label{fig1}
	\end{wrapfigure}
	We set up a two-level hierarchical RL structure similar to HAC. As shown in Figure 1, High level policy $\mu^{h}$ takes in state and outputs subgoals at intervals. Low level policy $\mu^{l}$ takes in state and desired subgoals to generate primitive actions. Here, low level policy has at most $K$ attempts of primitive action to achieve the desired subgoal, where $K$ can be viewed as the maximum horizon of a subgoal action is a hyperparameter given by the user. As long as the low level policy \(\mu^{l}\) run out of $K$ attempts or the desired subgoal is achieved, this high level transition terminates. The high level policy uses agent's current state as the new observation and produced another subgoal for low level policy to achieve. 
	
	We use an a binary reward function for low level policy learning in which a reward of 0 is granted only if the subgoal produced by high level policy is achieved and a reward of -1 otherwise. Note that the environment's return (i.e. whether the agent successfully accomplished the task) will not affect the reward received by the low level policy. In our evaluation on simulated robotics environments, we use the positional features of the observations as the representation for subgoals. A subgoal is judged to be achieved only if the distance between subgoal and the gripper's current position $s_{n+1}$ is less than threshold $l$.

	\subsection{Meta goal-generation for Hierarchical reinforcement learning}
	
	The primary motivation for our hierarchical meta reinforcement learning strategy is that, when people try to solve new tasks using prior experience, they usually focus on the overall strategy we used in previous tasks instead of the primitive action execution mechanism. For instance, when we try to use the knowledge learned from riding bicycle to accelerate learning for riding motorcycle, the primitive action execution mechanism is entirely different although they share a similar high-level strategy (e.g. learn how to keep balance first). Thus, we take advantage of our two-level hierarchy structure and propose a new meta-RL framework called meta goal-generation for hierarchical RL (MGHRL). Instead of learning to generate detailed strategy for new tasks, MGHRL learns to generate overall strategy (subgoals) given past experience and leaves the detailed method of how to achieve the subgoals for independent RL. We leverage PEARL framework~\citep{DBLP:conf/icml/RakellyZFLQ19} to independently train a high level meta-policy which is able to quickly adapt to new tasks and generate proper subgoals. Note that off-policy RL method is indispensable in our structure when training high level policy due to its excellent sample efficiency during meta-training. And its structured exploration by reasoning about uncertainty over tasks is crucial to hierarchical parallel training framework. We leave the low level policy to be trained independently with non-meta RL algorithm using hindsight experience replay mechanism~\citep{DBLP:conf/nips/AndrychowiczCRS17}. In our simulated robotics experiments, the low level policy aims to move the gripper to the desired subgoal position which can be reused when switching to other tasks. Thus we only need to train a single set of low-level polices which can be shared across different tasks. In other situations where the tasks are from different domains, we can choose to train low level policy independently on new tasks without using past experience. 
	
	We summarize our meta-training procedure in Algorithm~\ref{alg1}. For each training task drawn from task distribution, we sample context $c_{h}$ and generate hindsight transitions\footnote{To achieve parallel training for the two levels of our framework, we rewrite past experience transitions as hindsight action transitions, and supplement both levels with additional sets of transitions as was done in HAC.} for both levels of hierarchy ($line$ $4\sim13$) by performing current policy. Then we train high level and low level networks with the collected data ($line$ $16\sim22$).

		\begin{algorithm}[htb]
	    \caption{MGHRL Meta-training}
        \label{alg1}
        \begin{algorithmic}[1]
        \REQUIRE Batch of training tasks $\{\tau_{i}\}_{i=1,...,T}$ from $p(\tau)$, maximum horizon $K$ of subgoal action
        \STATE Initialize replay buffers $\mathcal{B}^{i}_{h}$,$\mathcal{B}^{i}_{l}$ for each training task
        \WHILE{not done}
        \FOR{each task $\tau_{i}$}
        \STATE Initialize high-level context $c_{h}^{i}=\{\}$
        \FOR{m=1,...,M}
        \STATE Sample $z\sim q_{\phi }(z|c^{i}_{h})$
        \STATE $g_{i}\leftarrow\mu_{h}(g|s,z)$
        \FOR{$K$ attempts or until $g_{i}$ achieved}
        \STATE Gather data using  $a_{i}\leftarrow\mu_{l}(a|s,g)$
        \STATE Generate hindsight action transition, hindsight goal transition and add to $\mathcal{B}^{i}_{l}$
        \ENDFOR
        \STATE Generate hindsight transitions, subgoal test transitions and add to $\mathcal{B}^{i}_{h}$
        \STATE Sample high level context $c_{h}^{i}=\{s_{j},g_{j},r_{j},s'_{j}\}_{j=1,...,N}\sim\mathcal{B}^{i}_{h}$ 
        \ENDFOR
        \ENDFOR
        \FOR{each training step}
        \FOR{each task $\tau_{i}$}
        \STATE Sample high level context $c_{h}^{i}\sim \mathcal{B}^{i}_{h}$ and RL batch $b^{i}_{h}\sim\mathcal{B}^{i}_{h}$, $b^{i}_{l}\sim\mathcal{B}^{i}_{l}$
        \STATE Sample $z\sim q_{\phi }(z|c^{i}_{h})$ and calculate $L_{actor}^{h}(b^{i}_{h},z)$, $L_{critic}^{h}(b^{i}_{h},z)$, $L_{KL}^{h}$
        \STATE Update low level actor and critic network with $b^{i}_{l}$
        \ENDFOR
        \STATE Update high level networks with $\sum_{i}L_{actor}^{h}$, $\sum_{i}L_{critic}^{h}$, $\sum_{i}L_{KL}^{h}$
        \ENDFOR
        \ENDWHILE
        \end{algorithmic}
	\end{algorithm}

	\section{Experiments}
	\label{sec4}
	We evaluated our algorithm on several challenging continuous control robotics tasks (integrated with OpenAI Gym), simulated via the MuJoCo physics simulator~\citep{DBLP:conf/iros/TodorovET12}:
	
	\textbf{Fetch-Reach} Fetch has to move the gripper to the desired goal position.
	
	\textbf{Fetch-Push} Fetch has to move a box by pushing it until it reaches a desired goal position.
	
	\textbf{Fetch-Slide} Fetch has to hit a puck across a long table such that it slides to rest on the desired goal.
	
	\textbf{Fetch-PickandPlace} Fetch has to pick up a box from a table using its gripper and move it to a desired goal located on the table. To make exploration easier we recorded a single state in which the gripper is a few distance right above the box and start the training episodes from this state.
	
	We compare our algorithm to baselines including PEARL with dense reward, HER-PEARL with sparse reward and HAC with shared policy. Note that Rakelly et al.~\citeyearpar{DBLP:conf/icml/RakellyZFLQ19} has already shown that PEARL greatly outperforms other existing meta-RL methods like MAML~\citep{DBLP:conf/icml/FinnAL17}, ProMP~\citep{DBLP:conf/iclr/RothfussLCAA19} at both sample efficiency and final performance. Thus we mainly compare our results with it. In sparse reward setting, we further modify PEARL with Hindsight Experience Replay~\citep{DBLP:conf/nips/AndrychowiczCRS17} for a fair comparison\footnote{We also evaluated PEARL (without HER) with sparse reward and it was not able to solve any of the tasks.}. The last one means we train a shared HAC policy jointly across all meta-train tasks sampled from the whole task distribution.
	
     We first do the simplest meta-learning evaluation on each type of the four tasks. In each scenario, we evaluate on 50 meta-train tasks and 10 meta-test tasks, where the difference between each task is in the terminal goal position we want the box or gripper to reach as well as the initial positions. The results are shown in Table~\ref{tab1}. In Fetch-reach environment which is very easy to learn as we mentioned before, the tested methods all reach a final performance of $100\%$ success rate. Our method MGHRL outperforms the other three methods in Push and Slide scenarios, while PEARL with dense reward performs better in Pick-Place tasks. Our two-level hierarchy and hindsight transitions significantly decrease the difficulty of meta learning with sparse reward, and is able to learn efficiently under a fixed budget of environment interactions. HAC with shared policy lacks generalization ability and cannot always achieve good performance when tested on varied tasks as shown in our results.  
	
	\begin{wraptable}{r}{9cm}
	\vspace{-0.3cm}
	\caption{Average success rates over all tasks (meta-test)}
	\label{tab1}
\begin{tabular}{lp{1.2cm}lll}
\hline
\bf Tasks & \bf MGHRL & \bf PEARL  &  \bf HER-PEARL & \bf HAC\\\hline 
\bf Reach & $100\%$ & $100\%$ & $100\%$ & $100\%$\\
\bf Push & $\textbf{76\%}$ & $61\%$ & $15\%$ & $41\%$\\
\bf Slide & $\textbf{36\%}$ & $5\%$ & $6\%$ & $23\%$\\
\bf Pick-Place & $92\%$ & $\textbf{98\%}$ & $47\%$ & $13\%$\\\hline
\vspace{-0.5cm}
\end{tabular}
\end{wraptable}
	
	We further evaluate our method on tasks with wider distribution. As shown in Figure~\ref{fig3}, each scenario's meta-train and meta-test tasks are sampled from the original two or three types of tasks (e.g. 30 meta-train tasks from Push and 30 meta-train tasks from Slide). Our algorithm MGHRL generally achieves better performance and adapts to new task much more quickly in all four types of scenarios. Directly using PEARL to learn a meta-policy that considers both overall strategy and detailed execution mechanism would decrease prediction accuracy and sample efficiency in these wider-distribution tasks as shown empirically. It is better to decompose the meta training process and focus on goal-generation learning. In this way, our agent only needs to learn a meta-policy that gives the learning rules for learning how to generate proper subgoals. Moreover, under dense reward settings of these challenging tasks, the critic of PEARL has to approximate a highly non-linear function that has different meanings for the two or three different types of tasks. Using the sparse return is much simpler since the critic only has to differentiate between successful and failed states.

	\begin{figure}[htbp]
		\centering
		\includegraphics[width=0.99\textwidth]{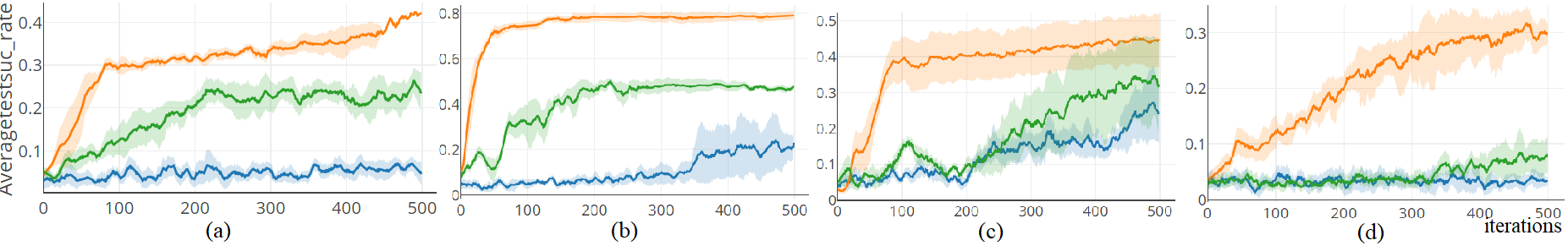}
	\end{figure}
	\begin{figure}[h]
		\centering
		\vspace{-0.5cm}
		\includegraphics[width=0.99\textwidth]{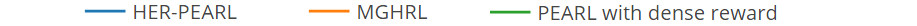}
		\caption{Average success rates for MGHRL, PEARL agents in each scenario: (a) Push $\&$ Slide $\&$ Pick-Place, (b) Push $\&$ Pick-Place, (c) Pick-Place $\&$ Slide, (d) Push $\&$ Slide. Each algorithm was trained for 1e6 steps. The error bar shows $1$ standard deviation.}
		\label{fig3}
	\end{figure}

	\section{Discussion and Future Work}
	In this paper, we propose a hierarchical meta-RL algorithm, MGHRL, which realizes meta goal-generation and leaves the low-level policy for independent RL. MGHRL focuses on learning the overall strategy of tasks instead of learning detailed action execution mechanism to improve the efficiency and generality. Our experiments show that MGHRL outperforms the SOTA especially in problems with relatively wider task distribution. Beyond this paper, we believe our algorithm can accelerate the acquisition of entirely new tasks. For example, to learn tasks such as riding bicycles and riding a motorcycle, the two primitive action execution mechanism are entirely different but the two learning process still share similar high-level structures (e.g. how to keep balance). With meta learning on high level policy, our algorithm is still supposed to achieve good performance on such tasks. We leave these for future work to explore.

\bibliography{iclr2020}
\bibliographystyle{iclr2020_conference}

\end{document}